\documentclass[sigconf]{acmart}

\usepackage{multirow}
\usepackage{makecell}
\usepackage{enumitem}

\newtheoremstyle{mystyle}
  {}
  {}
  {}
  {}
  {\bfseries}
  {.}
  { }
  {\thmname{#1}\thmnumber{ #2}\thmnote{ (#3)}}

\theoremstyle{mystyle}
\newtheorem{definition}{Definition}

\AtBeginDocument{%
  }

\setcopyright{none}
\renewcommand\footnotetextcopyrightpermission[1]{}
\acmConference[Conference acronym 'XX]{Make sure to enter the correct
  conference title from your rights confirmation email}{June 03--05,
  2018}{Woodstock, NY}
\acmISBN{978-1-4503-XXXX-X/18/06}

\begin{document}

\title{
MAPS: A Multimodal Agentic Framework for POI Data Production Based on Crowdsourced Street-view Imagery
}

\author{Sijie Ruan$^{1}$, Tingyu Xie$^{1,2\dagger}$, Mingfei Liu$^{2*}$, Yicheng Tang$^{1}$, Haoyu Han$^{1}$, Jingnan Qi$^{1}$,}
\author{Jingshu Zeng$^{3,2\dagger}$, Dongqi Xu$^{2}$, Jing Wang$^{2}$, Shuliang Wang$^{1*}$}
\affiliation{%
  \institution{$^1$Beijing Institute of Technology, Beijing \quad $^2$Kuaishou Technology, Beijing \\ $^3$The University of Hong Kong, Hong Kong SAR}
  \country{China}
}

\email{\{sjruan,xietingyu,tyc,hanhy,3220251485,slwang2011\}@bit.edu.cn}
\email{\{liumingfei,zengjingshu,xudongqi,wangjing64\}@kuaishou.com}

\renewcommand{\authors}{Sijie Ruan, Tingyu Xie, Mingfei Liu, Yicheng Tang, Haoyu Han, Jingnan Qi, Jingshu Zeng, Dongqi Xu, Jing Wang, Shuliang Wang}
\renewcommand{\shortauthors}{Sijie Ruan et al.}

\begin{abstract}
Point-of-interest (POI) databases support location-based services, yet urban changes quickly make their records stale and manual maintenance is costly. Crowdsourced street-view images provide timely visual evidence, but converting them into database actions requires separating multiple POIs and irrelevant text in each image, then resolving incomplete observations against ambiguous records. We study Street-view Image-based POI Data Production and propose MAPS, the first framework to integrate structured POI extraction with database maintenance from such imagery. MAPS contains two cascaded modules. SPICE progressively adapts a lightweight vision-language model to reconstruct and filter text, group entries belonging to the same POI, and classify their semantics and completeness. RIMA retrieves candidate records, performs evidence-grounded POI matching and attribute-level reasoning, and selectively acquires complementary visual, geospatial, or external-source evidence when necessary. In human-reviewed full-task evaluation, MAPS outperforms the strongest baseline by 21.8\%. MAPS has also been deployed in Kuaishou's large-scale POI production pipeline, reducing average processing time by 76.9\% relative to manual annotation. These results demonstrate the effectiveness and practical value of MAPS for scalable POI maintenance.
\end{abstract}



\newcommand{\SJ}{\textcolor{blue}}
\newcommand{\TBD}{\textcolor{red}}


\maketitle
\begingroup
  \renewcommand{\thefootnote}{\fnsymbol{footnote}}
  \footnotetext[2]{Work done during an internship at Kuaishou Technology.}
  \footnotetext[1]{Corresponding authors.}
\endgroup


\section{INTRODUCTION}
\label{sec:intro}

\begin{figure}[t]
  \centering
  \includegraphics[width=\columnwidth]{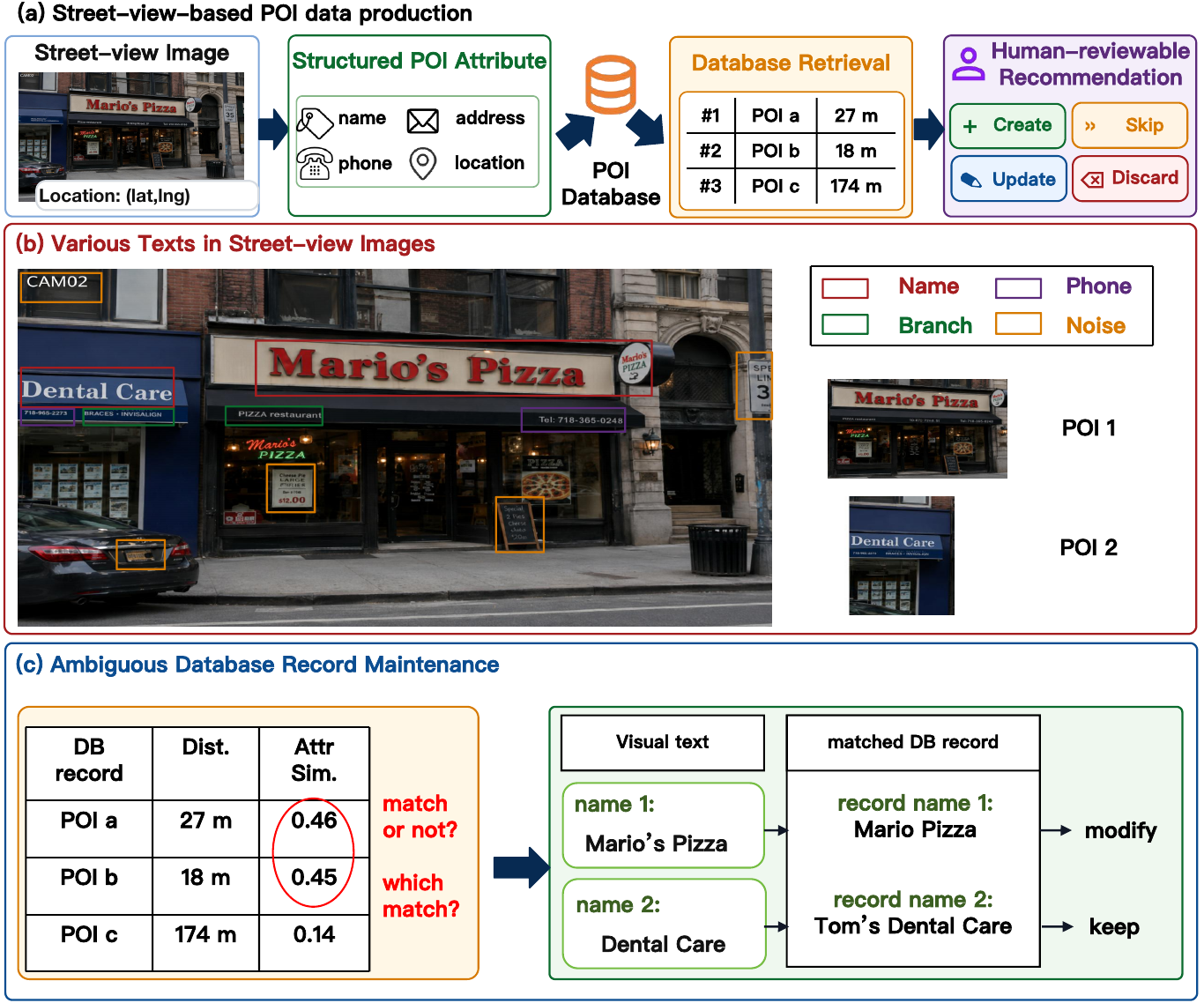}
  \caption{Street-view-based POI data production and its two challenges: various texts in street-view images, and ambiguous database record maintenance.}
  \Description{A three-panel illustration. Panel (a) converts a geotagged street-view image into structured POI attributes, retrieves nearby database records, and produces a human-reviewable create, update, skip, or discard recommendation. Panel (b) shows storefront text corresponding to POI names, branch names, phone numbers, and noise, with entries grouped into two POIs. Panel (c) shows retrieved records with similar distances and attribute scores, followed by decisions about which record matches and whether individual attributes should be modified or kept.}
  \label{fig:problem}
  \vspace{-10pt}
\end{figure}

Point-of-interest (POI) data provide essential infrastructure for location-based services, e.g., navigation~\cite{ruan2020doing}, location-based recommendation~\cite{lian2020geography}, credit services~\cite{han2019creditprint}. Nevertheless, due to the rapid development of cities, POIs, especially storefront POIs, may change from time to time, incurring the issue of data staleness, which severely degrades service accuracy and user experience. Conventional approaches for POI data production rely heavily on human-labors, e.g., field surveys, which are costly and inefficient. Volunteered Geographic Information (VGI)~\cite{goodchild2007citizens}, which infers geographic information based on crowdsourced spatio-temporal data, e.g., trajectories~\cite{ruan2020learning,ruan2020doing,ruan2022discovering}, geo-tagged images~\cite{poiCrowdsourcing2019}, provides an economical way to complement the geographic data production process. 
Widely accessible crowdsourced street-level imagery, captured by dashcams or handheld action cameras, offers a wealth of real-world street scenes, making it an ideal data source for updating POIs. Therefore, in this study, we investigate \emph{Street-view Image-based POI Data Production}, which generates a POI database action based on a given crowdsourced street-view image as shown in Figure~\ref{fig:problem}(a).

However, street-view image-based POI data production is non-trivial due to two major challenges.
\textit{Firstly}, there are various texts in street-view images, as shown in Figure~\ref{fig:problem}(b), which causes several issues: 
i) Some texts are not related to POIs, e.g., advertisements and traffic signs, which should be regarded as noises and filtered; 
ii) an image may contain the information of multiple POIs, the texts belonging to different POIs should be distinguished;
iii) some POIs have incomplete information because they are located at the edges of the street view images.
Existing methods commonly combine text recognized by OCR with visual and layout information, or use separate modules for text detection, recognition, and information fusion~\cite{m4c2020,latr2022,prestu2023,vlmc2022}. However, they do not jointly filter irrelevant text, group entries by POI, and identify attributes truncated at image boundaries.
\textit{Secondly}, after we extract the POI information from the image, leveraging them to maintain the POI database is still ambiguous, as shown in Figure~\ref{fig:problem}(c). It usually involves two steps: i) check whether the extracted POIs are in the database; ii) determine whether each attribute of the POIs should be updated or created. The former step is usually known as POI matching~\cite{sun2023conflating}. The de facto solution is to jointly consider both spatial proximity and attribute similarity to find the most similar POI in the database. However, due to the incompleteness of the observation or artistic typography design of storefronts, the correct POI may be failed to be retrieved. 
Even we successfully match the correct POI in the database, the extracted information may not always be used to update the attributes if the observation is incomplete or in lower quality.
Existing methods rely on fixed data sources and large amounts of labeled data, which limits their generalization~\cite{queryPoiMatching2020,dumapper2022,urbanPoi2023}. With incomplete observations or several nearby records with similar names, these fixed inputs may not identify the correct record, and the methods cannot actively obtain external information to resolve the ambiguity.

In recent years, multimodal vision-language models (VLMs) and agents have shown strong performance across diverse tasks by leveraging knowledge acquired during pre-training~\cite{qwen3_2025}, offering a promising foundation for street-view image-based POI data production. We therefore propose a \underline{M}ultimodal \underline{A}gentic framework for \underline{P}OI data production based on Crowdsourced \underline{S}treet-view Imagery (\textbf{MAPS}), comprising cascaded POI extraction and maintenance modules.
To tackle the challenge of \textit{Various Texts in Street-view Images}, the extraction module, \underline{S}treet-view-grounded \underline{P}OI \underline{I}nterpretation and \underline{C}omponent \underline{E}xtraction (\textbf{SPICE}), adapts a pre-trained lightweight vision-language model through a three-stage post-training process to filter noise, group information by POI, and classify entry semantics, thereby extracting structured POI information from a street-view image.
To tackle the challenge of \textit{Ambiguous Database Record Maintenance}, the maintenance module, \underline{R}etrieval-\underline{I}nformed POI \underline{M}aintenance \underline{A}gent (\textbf{RIMA}), treats top-ranked results from traditional POI matching methods as candidates and performs evidence-grounded POI matching and attribute-level reasoning. When the available evidence is insufficient or conflicting, it selectively acquires visual, geospatial, or external-source evidence for the corresponding matching or attribute-update decision.
Our contributions can be summarized as follows:

\vspace{-5pt}

\begin{itemize}[leftmargin=*]
  \item 
  To the authors' knowledge, we present the first end-to-end framework that integrates structured POI extraction, database maintenance from street-view imagery, establishing a practical paradigm for scalable crowdsourced image-based POI updating.
  \item 
  We propose MAPS with two cascaded modules: SPICE progressively transforms fragmented visual evidence into structured POI components; RIMA matches these components against database records, distinguishes existing from new POIs, and recommends evidence-supported attribute changes.
  \item 
  MAPS achieves an action accuracy of 84.8\% when deciding whether to match an observed POI to the correct existing record, create a new record, or discard the observation, representing a relative improvement of 21.8\% over the strongest baseline. It has also been deployed in Kuaishou's POI production pipeline, reducing average processing time by 76.9\% relative to manual annotation.
\end{itemize}

\section{PRELIMINARIES}
\label{sec:prelim}
\label{sec:formulation}

\begin{definition}[Point-of-interest]
A point-of-interest (POI), denoted by $p$, is a real-world place represented by the attribute tuple $(nm, br, loc, cat, addr, tel)$, where $nm$, $br$, $cat$, $addr$, and $tel$ denote the name, branch name (e.g., ``Times Square'' is the branch name in ``McDonald's (Times Square)''), category, address, and phone number, respectively. Its GPS location is $loc=(lat,lng)$, where $lat$ and $lng$ denote latitude and longitude.
\end{definition}

\begin{definition}[POI Database]
 A POI database stores currently known POIs, denoted as $\mathcal{B}=\{p_j\}_{j=1}^{N}$.
\end{definition}

\begin{definition}[Street-view Image] 
A street-view image is represented by $(I,\ell)$, where $I$ is the captured image and $\ell=(lat,lng)$ is its GPS location.
\end{definition}

We assume that camera heading and intrinsic parameters are unavailable and thus approximate each observed POI's location by the image capture coordinates, $loc \approx \ell$. When available, these parameters can support more accurate localization of individual POIs~\cite{hu2022beyond,zhanabatyrova2023automatic}.

\begin{definition}[POI Data Production Action]
By comparing with the POI database, each candidate POI in the captured image corresponds to a POI data production action $act=<decision, op>$, where $decision$ is the action type, and $op$ contains decision-dependent operations. $decision$ is chosen from following options:
\begin{itemize}[leftmargin=*]
    \item $\texttt{create}$, which means the candidate POI is not in the database, a new POI should be created. In this case, $op$ contains all information of the candidate POI to be created.
    \item $\texttt{update}$, which means the candidate POI already exists in the database, but one or more database attributes are outdated or previously missing information has become available. In this case, $op$ specifies only the attributes to be updated and their new values.
    \item $\texttt{skip}$, which also means the candidate POI is in the database, and all information of entry in the database is up-to-date. In this case, we skip this candidate POI, and $op$ is empty.
    \item $\texttt{discard}$, which means the candidate POI cannot be reliably determined and should therefore be omitted. In this case, $op$ is empty.
\end{itemize}
\end{definition}

\noindent\textbf{Problem Definition (Street-view Image-based POI Data Production).} 
\textit{
Given a street-view image $(I,\ell)$ and a POI database $\mathcal{B}$, generate a set $\mathcal{A}$ of POI data production actions, where each $act \in \mathcal{A}$ corresponds to a POI candidate observed in $I$.
}

Note that, in this study, we do not consider removing POIs from the database, since the absence of a POI from a single street-view image does not reliably indicate its closure, which should be regarded as as a different research problem that needs to jointly consider multiple street-view images at a location across different times.

\section{METHODOLOGY}
\label{sec:method}


In this section, we first outline the overall pipeline of MAPS and then detail SPICE for structured POI extraction and RIMA for evidence-grounded database maintenance.

\subsection{MAPS Overview}
\label{sec:maps_overview}

\begin{figure*}[!t]
  \centering
  \includegraphics[width=\textwidth]{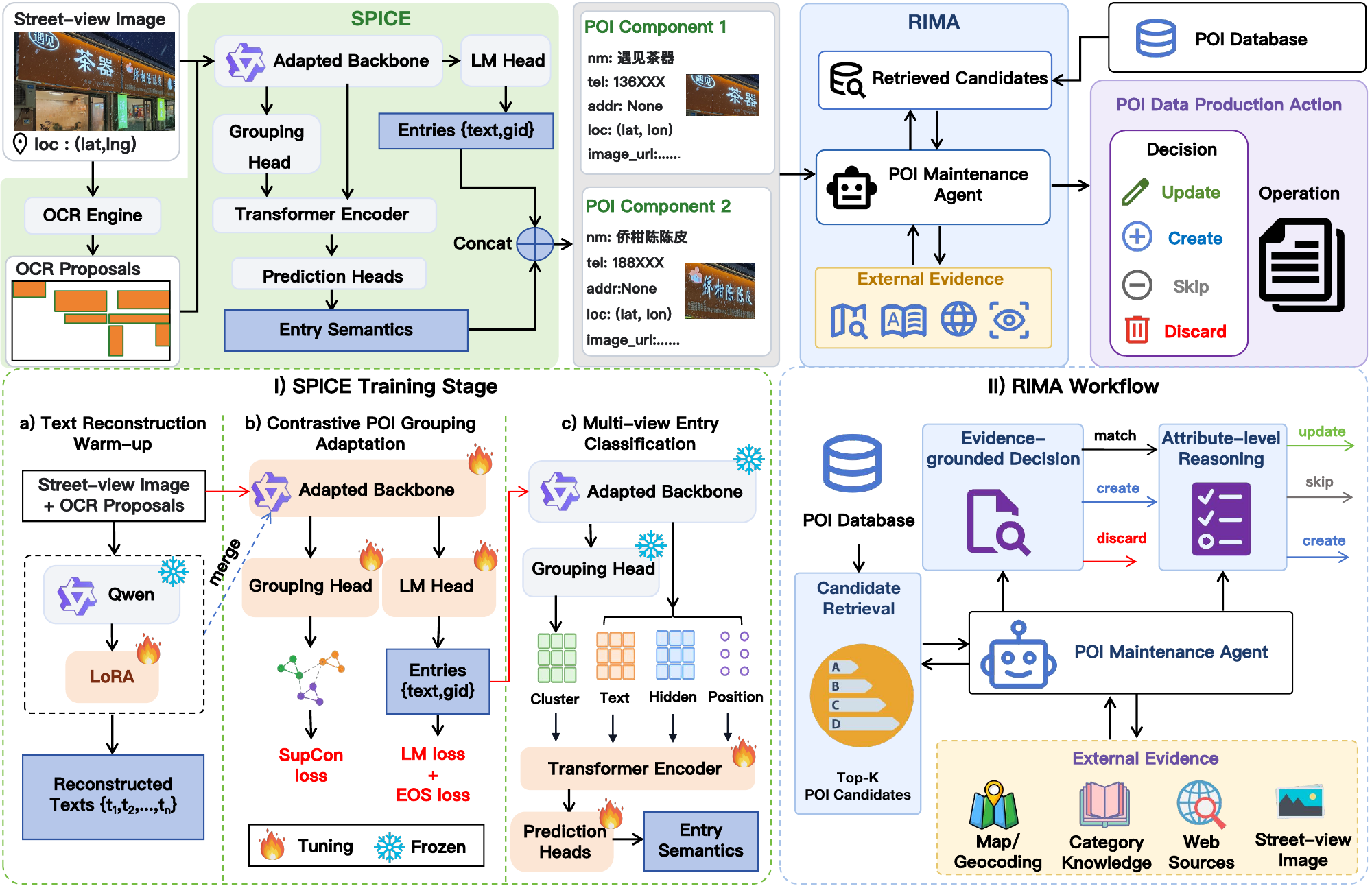}
  \caption{MAPS overview, including the POI data production pipeline, progressive SPICE training, and evidence-grounded RIMA workflow. Red arrows between SPICE stages denote stage inputs, solid black arrows show processing flows.}
  \Description{MAPS processes a street-view image with SPICE to produce structured POI components, which RIMA compares with retrieved database candidates and external evidence to generate POI data production actions. The detailed workflows show SPICE's three-stage training and RIMA's evidence-grounded decision and attribute-level reasoning.}
  \label{fig:overview}
  \vspace{-10pt}
\end{figure*}

Figure~\ref{fig:overview} depicts MAPS's POI data production in two steps:
\textbf{SPICE for POI Extraction} organizes visually observable POI information from $I$ into structured components while filtering unrelated text; each component retains $\ell$ as downstream evidence. \textbf{RIMA for POI Maintenance} compares each component with $\mathcal{B}$ and produces a POI data production $decision$ with its operations $op$, using the capture location primarily for candidate retrieval. The lower panels detail SPICE training and RIMA's decision process.

\vspace{-10pt}

\subsection{SPICE: Street-view-grounded POI Interpretation and Component Extraction}
\label{sec:spice}

Inspired by curriculum learning~\cite{curriculum2009}, SPICE addresses the various-text challenge through three progressive stages. \textit{Text Reconstruction Warm-up} first uses OCR proposals and a compact vision-language backbone to correct and recover visually supported text~\cite{m4c2020,latr2022,prestu2023}. \textit{Contrastive POI Grouping Adaptation} then generates group identifiers under contrastive regularization to separate entries belonging to different POIs. Finally, \textit{Multi-view Entry Classification} combines visual-language, grouping, textual, and spatial information to predict each entry's validity, semantic role, and completeness, filtering unrelated text and assembling the retained entries into structured POI components.

\subsubsection{Text Reconstruction Warm-up}
\label{sec:vtu}

An OCR engine~\cite{paddleocr2020,ppocrv5_2026,vlmc2022} first returns auxiliary proposals $D=\{(\tilde{t}_j,\beta_j)\}_{j=1}^{J}$, where $\tilde{t}_j$ and $\beta_j$ are the preliminary transcription and detected region of proposal $j$. Given $I$ and $D$, the model re-examines the visual regions to correct recognition errors, merge fragmented proposals, and recover visually supported text missed by OCR. The proposal-to-entry association is not restricted to one-to-one; detailed association and box construction are provided in Appendix~\ref{app:spice_details}.

We first use LoRA~\cite{lora2022} to adapt a Qwen vision-language backbone~\cite{qwen3_2025} for autoregressive generation of corrected text entries from $I$ and $D$. This stage focuses only on visual-text reconstruction and does not predict POI groups or semantic roles. In the next stage, we continue training the same backbone to learn POI grouping.

\vspace{-10pt}

\subsubsection{Contrastive POI Grouping Adaptation}
\label{sec:spg}

This stage continues training the backbone from Text Reconstruction Warm-up, with all backbone parameters unfrozen. Given $I$ and $D$, the LM head now generates both reconstructed text and an explicit POI group identifier. For entry $i$, its output is $\hat e_i=(\hat t_i,\hat g_i)$, where $\hat t_i$ is the reconstructed text and $\hat g_i$ is the generated group identifier. Entries with the same positive $\hat g_i$ are assigned to the same perceived POI, while $\hat g_i=0$ indicates that the entry is not assigned to a POI group.


Alongside the LM head, a lightweight grouping head maps the final hidden state of entry $i$ to an $L_2$-normalized representation $h_i^{\mathrm{grp}}\in\mathbb{R}^{d_{\mathrm{grp}}}$, where $d_{\mathrm{grp}}$ is the representation dimension.

We jointly optimize structured generation, POI grouping, and output termination:
\begin{equation}
  \mathcal{L}_{\mathrm{grp}}
  =
  \mathcal{L}_{\mathrm{LM}}
  +\lambda_g\mathcal{L}_{\mathrm{SupCon}}
  +\lambda_e\mathcal{L}_{\mathrm{EOS}}.
  \label{eq:spg_loss}
\end{equation}
Here, $\mathcal{L}_{\mathrm{grp}}$ is the total loss for this stage. $\mathcal{L}_{\mathrm{LM}}$ supervises the generation of reconstructed-text and group-identifier; $\mathcal{L}_{\mathrm{SupCon}}$ is the supervised contrastive loss for learning POI grouping~\cite{supcon2020}; and $\mathcal{L}_{\mathrm{EOS}}$ supervises correct output termination. The coefficients $\lambda_g$ and $\lambda_e$ weight the contrastive and termination losses, respectively. The contrastive loss is constructed within each image: entries from the same POI are positive pairs, entries from different POIs are negative pairs, and entries with group identifier zero are excluded. Its full formulation is provided in Appendix~\ref{app:spice_details}.

At inference, the LM head directly generates $\hat g_i$, so entries sharing a positive identifier are grouped without an additional clustering algorithm. The contrastive loss is used during training to learn a more discriminative $h_i^{\mathrm{grp}}$, and the grouping head is retained at inference to provide this representation as the cluster view for Multi-view Entry Classification.

\subsubsection{Multi-view Entry Classification}
\label{sec:cls}

For each generated entry $i$, the classifier combines four complementary views. The hidden view $h_i^{\mathrm{hid}}$ captures the image and generated-text context. The grouping view $h_i^{\mathrm{grp}}$ represents the POI membership learned in the previous stage. The text view $h_i^{\mathrm{txt}}$ captures the lexical meaning of the reconstructed text, while the position view $h_i^{\mathrm{pos}}$ describes the location and shape of the entry region in the image.

Each view is projected separately, and the projected views of entry $i$ are concatenated into $u_i$. A Transformer encoder~\cite{vaswani2017attention} then contextualizes all entries in the same image:
\begin{equation}
\begin{aligned}
  u_i=[\phi_h(h_i^{\mathrm{hid}});\phi_g(h_i^{\mathrm{grp}});
         \ \phi_t(h_i^{\mathrm{txt}});\phi_p(h_i^{\mathrm{pos}})],\\
  (H_1,\ldots,H_{N_e})={}&\operatorname{TF}(u_1,\ldots,u_{N_e}),
\end{aligned}
  \label{eq:fusion}
\end{equation}
Here, $[\,;\,]$ denotes concatenation. The functions $\phi_h$, $\phi_g$, $\phi_t$, and $\phi_p$ are the learned projections for the hidden, grouping, text, and position views, respectively. $N_e$ is the number of generated entries in the image, $\operatorname{TF}$ is the Transformer encoder, and $H_i$ is its contextualized representation for entry $i$. Three prediction heads map $H_i$ to validity $\hat v_i$, semantic role $\hat r_i$, and completeness $\hat q_i$. The semantic-role output is
\begin{equation}
  \hat r_i\in\{nm, br, addr, tel, \texttt{other}\},
  \label{eq:role_set}
\end{equation}
Here, $\hat v_i$ indicates whether entry $i$ contains usable POI information, $\hat q_i$ whether its visible text is complete, and the five values above are the possible roles of $\hat r_i$. A weighted sum of the three losses trains the projections, Transformer encoder, and prediction heads, while the Stage-2 backbone and grouping head remain frozen. Detailed view construction and the full objective are provided in Appendix~\ref{app:classifier_details}.

At inference, invalid entries are removed, and the remaining valid entries are assembled into components according to their predicted POI groups. Name and branch entries define the component name, address and phone entries populate its attributes, and \texttt{other} entries remain auxiliary evidence. The completeness prediction identifies truncated attributes. The resulting $m$-th component is $c_m=(nm_m,A_m,I,\ell)$, where $nm_m$ is its name, $A_m$ its extracted attributes, and $(I,\ell)$ the supporting image and capture location passed to RIMA.

\subsection{RIMA: Retrieval-Informed POI Maintenance Agent}
\label{sec:rima}
RIMA addresses ambiguous database record maintenance in three steps. \textit{Candidate Retrieval} first narrows $\mathcal{B}$ to nearby top-$K$ records without treating retrieval rank as entity identity. RIMA then performs \textit{Evidence-grounded Decision}, comparing the component with these records and selectively gathering auxiliary evidence to determine whether it matches an existing POI, provides sufficient evidence to create a new one, or should be discarded. Finally, in \textit{Attribute-level Reasoning}, it validates individual attributes and converts the entity state into the final $decision$ and its decision dependent $op$. This ordering prevents an uncertain identity decision from propagating to attribute maintenance.

\subsubsection{Candidate Retrieval}
\label{sec:retrieval}
Candidate Retrieval first queries the POI database $\mathcal{B}$ for records within a spatial range around the capture location. It then applies a neural matching model to rank these local records using name, address, phone, and location similarities~\cite{devlin2019bert,reimers2019sentence}. This model is an established module of Kuaishou's production pipeline for POI candidate retrieval and scoring. It returns
\begin{equation}
\label{eq:candidate_retrieval}
\mathcal{R}_K(c_m,\mathcal{B})=\{p_{m,1},\ldots,p_{m,K}\},
\end{equation}
where $p_{m,k}\in\mathcal{B}$ is the candidate at rank $k$. The resulting score is used only to select and rank the top-$K$ candidates and does not establish entity identity.

\subsubsection{Evidence-grounded Decision}
\label{sec:judge}
In Evidence-grounded Decision, RIMA considers $c_m$, the ranked candidates, and visual, spatial, and database evidence to produce
\begin{equation}
  s_m\in\{\texttt{match},\texttt{create},\texttt{discard}\},
  \label{eq:judge_output}
\end{equation}
where \texttt{match} additionally identifies a record $p_m^*\in\mathcal{R}_K(c_m,\mathcal{B})$. Here, \texttt{match} is an internal entity state: it is later converted to the final decision \texttt{update} or \texttt{skip}.

Retrieval rank is supporting evidence rather than a hard decision rule, so a lower-ranked candidate may be selected when its combined evidence is more consistent. The agent returns \texttt{create} only when the component is visually supported but no candidate represents the same entity, and \texttt{discard} when the evidence supports neither matching nor creation.

When initial evidence is insufficient or conflicting, RIMA selectively acquires map or geocoding evidence, category knowledge, web evidence, or candidate-conditioned visual verification from the street-view image; detailed triggers are given in Appendix~\ref{app:rima_details}.

\subsubsection{Attribute-level Reasoning}
\label{sec:attribute}

In Attribute-level Reasoning, RIMA evaluates the name, branch name, location, category, address, and phone attributes using the component, optional matched record, and collected evidence. Each attribute is assessed independently to determine whether the evidence supports adding or changing its value, retaining the database value, or omitting an unreliable observation. For a \texttt{match}, at least one supported attribute change yields \texttt{update}; otherwise the result is \texttt{skip}. For \texttt{create}, only validated attributes are retained, while \texttt{discard} produces no operation. The attribute decisions and supporting evidence are retained for human review. Detailed rules and the output schema are provided in Appendix~\ref{app:attribute_details}.

\section{EXPERIMENTS}
\label{sec:exp}

We evaluate MAPS through five research questions:
\begin{itemize}[leftmargin=*]
  \item \textbf{RQ1 (POI extraction, target-domain):} Does SPICE improve POI extraction in target domains?
  \item \textbf{RQ2 (POI extraction, zero-shot):} How does SPICE generalize to external street-view benchmark data in a zero-shot setting?
  \item \textbf{RQ3 (POI extraction, ablation):} What are the contributions of individual SPICE design modules?
  \item \textbf{RQ4 (POI maintenance):} How well does RIMA maintain POIs from manually verified components?
  \item \textbf{RQ5 (Full-task POI data production):} Does MAPS produce human-acceptable and practically useful recommendations from raw images?
\end{itemize}

\subsection{Experimental Setup}
\label{sec:setup}

\subsubsection{Datasets}
\label{sec:datasets}

Table~\ref{tab:datasets} summarizes all data splits. For POI extraction evaluation, the target-domain dataset combines vehicle-mounted street-view images from Kuaishou Dashcam with storefront-oriented images from the public Meituan ReCTS dataset. Kuaishou Dashcam captures blur, occlusion, perspective distortion, and distant text under real driving conditions, whereas Meituan ReCTS has denser, more regular storefront-centered layouts. MAPbench provides the external zero-shot extraction set; it is sourced from Amap public data and entirely excluded from SPICE training and validation. We normalize its annotations into our POI-entry and group representation (Appendix~\ref{sec:appendix_mapbench}).

\begin{table}[h]
  \caption{Dataset summary. Retrieval and maintenance use the Kuaishou POI database snapshot dated 2026-04-01; pos./neg. denote images with/without POIs, and U/C/S denote update/create/skip.}
  \label{tab:datasets}
  \centering
  \footnotesize
  \setlength{\tabcolsep}{2pt}
  \begin{tabular}{@{}
    >{\raggedright\arraybackslash}p{0.15\columnwidth}
    >{\raggedright\arraybackslash}p{0.11\columnwidth}
    >{\raggedright\arraybackslash}p{0.26\columnwidth}
    >{\raggedleft\arraybackslash}p{0.12\columnwidth}
    >{\raggedright\arraybackslash}p{0.26\columnwidth}@{}}
    \toprule
    \textbf{Evaluation} & \textbf{Split} & \textbf{Source} & \textbf{Size} & \textbf{Composition} \\
    \midrule
    \multirow{4}{0.15\columnwidth}{POI Extraction} & Train & Kuaishou Dashcam + Meituan ReCTS~\cite{rects2019} & 33,044 img. & 22,684 pos. / 10,360 neg. \\
    & Target test & Kuaishou Dashcam & 1,520 img. & 596 pos. / 924 neg. \\
    & Target test & Meituan ReCTS & 1,593 img. & 1,541 pos. / 52 neg. \\
    & Zero-shot test & MAPbench~\cite{mapbench2026} & 475 img. & Held out from training \\
    \midrule
    \multirow{2}{0.15\columnwidth}{POI Maintenance} & Test & Kuaishou Dashcam & 200 comp. & 150 U / 28 C / 22 S \\
    & Test & MAPbench & 241 comp. & 138 U / 54 C / 49 S \\
    \midrule
    POI Data Production & Review & Sampled from Kuaishou Dashcam & 197 img. & 315 component anchors \\
    \bottomrule
  \end{tabular}
\end{table}

For POI maintenance evaluation, we use the 441 manually verified Kuaishou Dashcam and MAPbench components in Table~\ref{tab:datasets}. Each visually supported component is paired with image evidence and retrieved candidates, removing upstream extraction errors to isolate RIMA. Reviewers label identity as \texttt{match} or \texttt{create}, matched components as \texttt{update}/\texttt{skip} with affected attributes. Ambiguous or low-quality components are excluded; hence this setting has no ground-truth \texttt{discard} cases, which are evaluated only in the raw-image full-task experiment. Appendix~\ref{sec:appendix_human_review} gives the review protocol.

For raw-image full-task evaluation, each pipeline starts from the same 197 images and is assessed on its own outputs against 315 common human-reviewed anchors. Unlike the component-level setting, it retains upstream extraction errors, allowing unreliable outputs to be discarded and the complete MAPS pipeline to be evaluated. Because extraction modules may produce different component sets, we measure recommendation acceptability and maintenance yield rather than city-level discovery recall. Appendices~\ref{sec:appendix_human_review} and~\ref{sec:appendix_full_task_protocol} describe the protocols.

\begin{table*}[!t]
  \caption{Target-domain POI extraction on Kuaishou Dashcam and Meituan ReCTS. Best and second-best F1 scores are bolded and underlined, respectively; time is measured on the combined test sets, and $\dagger$ denotes remote-API latency.}
  \label{tab:target_domain}
  \centering
  \resizebox{\textwidth}{!}{%
  \begin{tabular}{ll c ccccc ccccc c}
    \toprule
    \textbf{Cat.} & \textbf{Method} & \textbf{Dep.} &
    \multicolumn{5}{c}{\textbf{Kuaishou Dashcam}} &
    \multicolumn{5}{c}{\textbf{Meituan ReCTS}} &
    \makecell{\textbf{Time}\\\textbf{(combined)}} \\
    \cmidrule(lr){4-8} \cmidrule(lr){9-13}
    & & & Avg. F1 & $F1_{\mathrm{hp}}$ & $F1_{\mathrm{nm}}$ & $F1_{\mathrm{txt}}$ & $F1_{\mathrm{attr}}$
    & Avg. F1 & $F1_{\mathrm{hp}}$ & $F1_{\mathrm{nm}}$ & $F1_{\mathrm{txt}}$ & $F1_{\mathrm{attr}}$
    & (s/img) \\
    \midrule
    Traditional & VLMC & Local & 0.462 & 0.423 & 0.327 & 0.382 & 0.717 & 0.645 & 0.871 & 0.547 & 0.360 & 0.800 & 0.40 \\
    \midrule
    \multirow{4}{*}{Local}
    & Qwen3.5-0.8B & Local & 0.513 & 0.583 & 0.450 & 0.415 & 0.604 & 0.612 & 0.960 & 0.576 & 0.269 & 0.643 & 11.14 \\
    & Qwen3-VL-8B & Local & 0.717 & 0.762 & \underline{0.733} & 0.578 & 0.795 & 0.736 & 0.983 & \underline{0.708} & 0.429 & 0.826 & 8.94 \\
    & Qwen3.5-9B & Local & 0.719 & 0.764 & 0.707 & 0.668 & 0.738 & 0.717 & 0.967 & 0.680 & 0.407 & 0.815 & 14.04 \\
    & Qwen3.5-4B & Local & 0.712 & \underline{0.833} & 0.729 & \underline{0.670} & 0.615 & 0.686 & 0.965 & 0.636 & 0.375 & 0.768 & 13.71 \\
    \midrule
    \multirow{5}{*}{API}
    & Qwen3-VL-32B & API & 0.698 & 0.829 & 0.669 & 0.617 & 0.677 & 0.731 & 0.983 & 0.676 & 0.403 & 0.863 & 7.01$^{\dagger}$ \\
    & GPT-4.1 & API & 0.528 & 0.720 & 0.297 & 0.497 & 0.599 & 0.700 & \textbf{0.990} & 0.579 & 0.366 & 0.865 & 8.14$^{\dagger}$ \\
    & Qwen3.5-35B & API & 0.723 & 0.782 & 0.672 & 0.605 & \textbf{0.830} & 0.744 & 0.984 & 0.666 & 0.451 & \underline{0.874} & 4.66$^{\dagger}$ \\
    & Qwen3.5-397B & API & 0.740 & 0.801 & 0.701 & 0.652 & 0.807 & \underline{0.752} & 0.983 & 0.702 & \underline{0.456} & 0.866 & 8.53$^{\dagger}$ \\
    & Qwen3.5-Plus & API & \underline{0.743} & 0.802 & 0.711 & 0.648 & \underline{0.813} & 0.751 & 0.981 & 0.707 & 0.455 & 0.860 & 9.66$^{\dagger}$ \\
    \midrule
    \textbf{Ours} & \textbf{SPICE} & \textbf{Local} & \textbf{0.762} & \textbf{0.860} & \textbf{0.741} & \textbf{0.721} & 0.727 & \textbf{0.794} & \underline{0.989} & \textbf{0.717} & \textbf{0.558} & \textbf{0.911} & 3.91 \\
    \bottomrule
  \end{tabular}%
  }
\end{table*}

\subsubsection{Evaluation Metrics}
\label{sec:metrics}
We evaluate MAPS at three scopes: POI extraction, POI maintenance, and full-task POI data production.

\noindent\textbf{POI extraction metrics.}
We evaluate the successive outputs of SPICE. $F1_{\mathrm{hp}}$ measures whether the model correctly identifies images containing at least one POI. $F1_{\mathrm{nm}}$ evaluates the extracted POI names. $F1_{\mathrm{txt}}$ measures text extraction after predicted and ground-truth POI groups are aligned, so text assigned to the wrong POI is penalized. $F1_{\mathrm{attr}}$ is the macro F1 over the five entry roles in Eq.~\eqref{eq:role_set}: name, branch name, address, phone number, and \texttt{other}. The \emph{Average F1 Score of POI Extraction Task} (Average F1) is the equal-weight average of these four metrics, and time is reported per image. Detailed text matching, group alignment, and counting rules are provided in Appendix~\ref{sec:appendix_extraction_metrics}.

\noindent\textbf{POI maintenance metrics.}
We evaluate the three levels of RIMA. $L1$ measures whether the component is correctly routed to \texttt{match} or \texttt{create}; a matched prediction is correct only when it selects the correct database record. $L2$ evaluates the subsequent \texttt{update}/\texttt{skip} outcome for matched components. $L3$ evaluates attribute-level updates for name, branch name, location, category, address, and phone number. Errors at $L1$ propagate to $L2$ and $L3$, reflecting the complete maintenance chain. The \emph{Average F1 Score of POI Maintenance Task} (Average F1) averages the three levels; we also report LLM calls and time per component. Full definitions are provided in Appendix~\ref{sec:appendix_maintenance_metrics}.

\noindent\textbf{Full-task metrics.}
Action Accuracy evaluates \texttt{create}, \texttt{match}, and \texttt{discard} decisions and requires the correct record for a match. Attribute Accuracy evaluates \texttt{update}/\texttt{skip} decisions and attribute changes. Accepted Recommendations per 100 Images measures yield, and Completion normalizes it by the reviewed upper bound. We also report image-to-recommendation latency. Appendices~\ref{sec:appendix_full_task_protocol} and~\ref{sec:appendix_human_review} give the full protocol and review criteria.

\subsubsection{Baselines}
\label{sec:baselines}
Accordingly, we select baselines for POI extraction, POI maintenance, and their full-task combinations.

\noindent\textbf{POI extraction baselines.}
We compare SPICE with a traditional POI extraction pipeline and generic VLMs deployed either locally or through APIs, all under the same structured output definition. Most traditional approaches address isolated subtasks such as text detection, recognition, or POI classification and do not produce the entry-level semantics and POI-group structure required by our evaluation. Among them, VLMC~\cite{vlmc2022} is the only detection--recognition--fusion pipeline that can be adapted to this unified output. The VLM baselines comprise Qwen-family models~\cite{qwen3_2025} and GPT-4.1~\cite{gpt4_2025}; all are evaluated zero-shot using the same prompt.

\noindent\textbf{POI maintenance baselines.}
All methods receive the same verified POI component and access to the same POI database snapshot. Random samples routing and attribute decisions according to the empirical label distributions. Following the established taxonomy of POI conflation methods~\cite{sun2023conflating}, Rule-based retrieves nearby records, ranks them using weighted name and distance similarities, and applies deterministic rules to entity matching and attribute updates. ML-based relies throughout on compact BERT models~\cite{devlin2019bert}: it performs entity matching using a neural matching score and attribute comparison using attribute-specific models. These non-LLM methods use structured attributes but no image or external evidence. For the VLM-agent comparison, One-Shot, ReAct, and RIMA receive the same top-$K$ candidates from the neural matching model and have access to the same visual and auxiliary evidence sources. One-Shot produces entity and attribute decisions in a single prompt. ReAct~\cite{react2022} freely plans how to use the available evidence and tools, while RIMA follows the staged decision-before-attribute workflow.

\noindent\textbf{Full-task baselines.}
We form a $2\times2$ comparison by pairing two extraction methods, zero-shot Qwen3.5-Plus and SPICE, with two maintenance methods, ML-based and RIMA. This comparison separates the effects of POI extraction, POI maintenance, and their integration.

\subsubsection{Implementation Settings}
\label{sec:implementation_settings}

SPICE uses Qwen3.5-0.8B and is trained and evaluated locally on four 48GB GPUs. RIMA and the VLM-agent baselines are executed through model APIs with temperature~0. All retrieval-based methods use $K{=}5$ candidates; Appendix~\ref{sec:appendix_hyperparams} gives the full training and inference settings.

\subsection{POI Extraction Evaluation}
\label{sec:poi_extraction_eval}

\subsubsection{Target-domain POI Extraction}\label{sec:target_domain} As shown in Table~\ref{tab:target_domain}, SPICE achieves the best Average F1 on both target-domain datasets. Its substantial improvement over the traditional VLMC pipeline shows that SPICE can more effectively extract structured POI records directly from street-view images. It also outperforms the strongest generic VLM by 1.9 and 4.2 Average-F1 points and ranks first on six of the eight component metrics, indicating consistent gains across the evaluated extraction subtasks. Moreover, SPICE runs at 3.91\,s/image, 2.3$\times$ faster than the fastest local VLM baseline.

\begin{figure}[!t]
  \centering
  \includegraphics[width=\columnwidth]{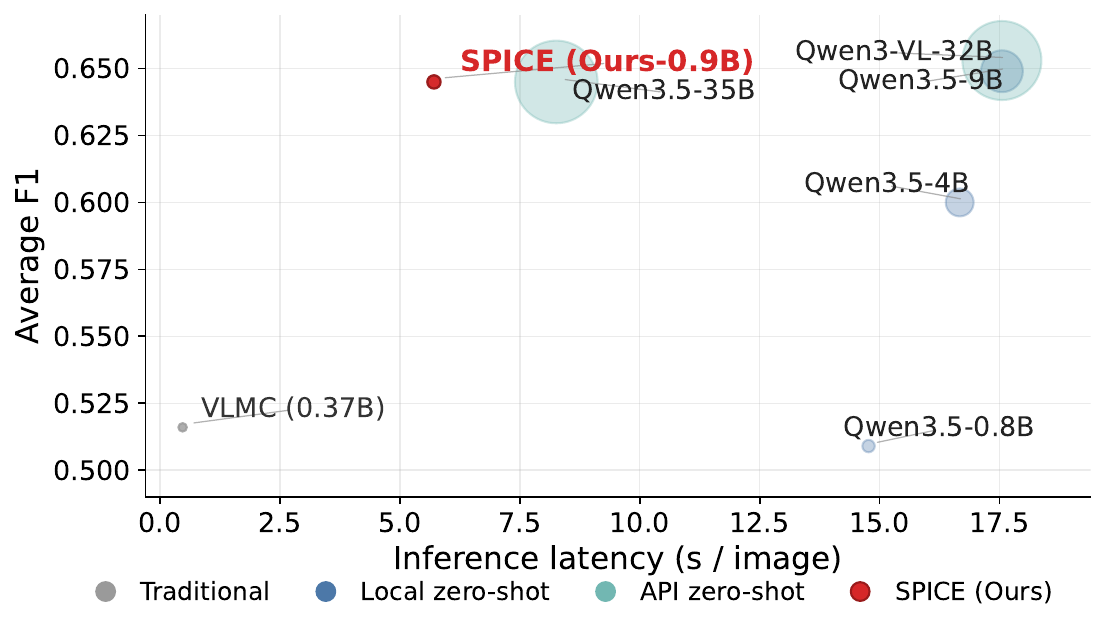}
  \caption{Zero-shot POI extraction on MAPbench: Average F1 versus latency. Bubble size indicates model scale; colors denote method categories.}
  \Description{Scatter plot showing the Average F1 of POI extraction on MAPbench against inference latency.}
  \label{fig:zero_shot}
\end{figure}

\subsubsection{External Zero-shot Generalization}\label{sec:zero_shot} Figure~\ref{fig:zero_shot} further evaluates SPICE on MAPbench, which is entirely excluded from training and validation. SPICE attains an Average F1 of 0.645 with approximately 0.9B parameters and 5.71\,s/image latency, remaining competitive with zero-shot VLMs that are one to two orders of magnitude larger. SPICE also clearly outperforms direct zero-shot prompting of its Qwen3.5-0.8B backbone. Since both use the same backbone, this gain demonstrates that the extraction, grouping, and classification capabilities learned by SPICE generalize to the zero-shot domain.

\begin{table*}[t]
  \caption{POI maintenance on randomly sampled, manually verified components from Kuaishou Dashcam and MAPbench. Best and second-best F1 scores are bolded and underlined, respectively; LLM calls and time are reported per component.}
  \label{tab:poi_maintenance}
  \centering
  \resizebox{\textwidth}{!}{%
  \begin{tabular}{ll l cccc cccc c c}
    \toprule
    & & & \multicolumn{4}{c}{\textbf{Kuaishou Dashcam}} & \multicolumn{4}{c}{\textbf{MAPbench}} & & \\
    \textbf{Cat.} & \textbf{Method} & \textbf{Backbone} &
    Avg. F1 & $L1$-F1 & $L2$-F1 & $L3$-F1 &
    Avg. F1 & $L1$-F1 & $L2$-F1 & $L3$-F1 &
    \textbf{LLM Calls} & \textbf{Time} \\
    \cmidrule(lr){4-7} \cmidrule(lr){8-11}
    & & & & & & & & & & & & (s/comp.) \\
    \midrule
    \multirow{3}{*}{\makecell{Non-LLM\\Baselines}}
    & Random & -- & 0.124 & 0.172 & 0.156 & 0.043 & 0.167 & 0.209 & 0.248 & 0.043 & 0 & -- \\
    & Rule-based & -- & 0.183 & 0.234 & 0.189 & 0.126 & 0.231 & 0.253 & 0.306 & 0.135 & 0 & -- \\
    & ML-based & -- & 0.363 & 0.476 & 0.275 & 0.338 & 0.535 & 0.602 & 0.621 & 0.383 & 0 & -- \\
    \midrule
    \multirow{12}{*}{VLM Agent}
    & \multirow{4}{*}{One-Shot} & Qwen3-VL-8B & 0.497 & 0.673 & 0.518 & 0.300 & 0.450 & 0.626 & 0.466 & 0.258 & 1 & 5.68 \\
    & & Qwen3.6-35B & 0.610 & 0.659 & 0.570 & 0.601 & 0.640 & 0.804 & 0.567 & 0.548 & 1 & 2.89 \\
    & & GPT-4.1 & 0.624 & 0.628 & 0.598 & 0.645 & 0.697 & 0.826 & 0.740 & 0.527 & 1 & 5.58 \\
    & & Qwen3.5-Plus & 0.752 & 0.769 & 0.695 & 0.792 & 0.759 & \underline{0.872} & 0.697 & 0.707 & 1 & 6.82 \\
    \cmidrule(lr){2-13}
    & \multirow{4}{*}{ReAct} & Qwen3-VL-8B & 0.658 & 0.695 & 0.641 & 0.636 & 0.675 & 0.729 & 0.751 & 0.544 & 10.7 & 25.3 \\
    & & Qwen3.6-35B & 0.686 & 0.783 & 0.723 & 0.553 & 0.673 & 0.761 & 0.742 & 0.516 & 9.5 & 21.6 \\
    & & GPT-4.1 & 0.578 & 0.571 & 0.537 & 0.625 & 0.599 & 0.626 & 0.607 & 0.563 & 7.2 & 19.8 \\
    & & Qwen3.5-Plus & 0.706 & 0.720 & 0.677 & 0.721 & 0.784 & 0.855 & 0.834 & 0.663 & 8.8 & 43.7 \\
    \cmidrule(lr){2-13}
    & \multirow{4}{*}{\textbf{RIMA (Ours)}} & \textbf{Qwen3-VL-8B} & 0.685 & 0.707 & 0.665 & 0.683 & 0.811 & 0.776 & 0.874 & 0.785 & 6.1 & 8.98 \\
    & & \textbf{Qwen3.6-35B} & \underline{0.849} & \textbf{0.860} & \underline{0.855} & 0.832 & 0.856 & 0.853 & \underline{0.884} & 0.832 & 6.2 & 7.13 \\
    & & \textbf{GPT-4.1} & 0.814 & \underline{0.810} & 0.785 & \textbf{0.848} & \underline{0.867} & 0.852 & 0.880 & \underline{0.868} & 6.2 & 13.1 \\
    & & \textbf{Qwen3.5-Plus} & \textbf{0.858} & \textbf{0.860} & \textbf{0.879} & \underline{0.836} & \textbf{0.880} & \textbf{0.873} & \textbf{0.896} & \textbf{0.872} & 6.3 & 14.8 \\
    \bottomrule
  \end{tabular}%
  }
\end{table*}

\subsubsection{Ablation Study of SPICE}\label{sec:ablation} 
\begin{figure}[t]
  \centering
  \includegraphics[width=\columnwidth]{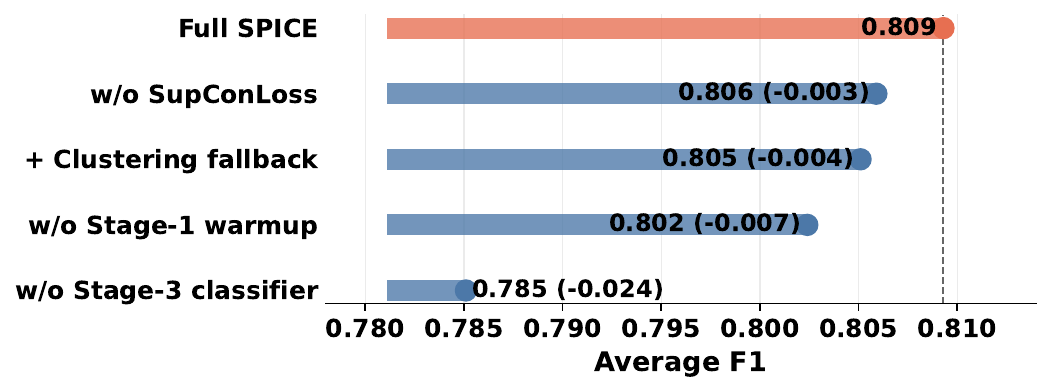}
  \caption{SPICE Average F1 ablation on the combined target-domain test sets; parentheses show changes from Full SPICE.}
  \Description{Horizontal ablation plot comparing the Average F1 of Full SPICE with variants that remove supervised contrastive loss, add clustering fallback, remove Stage-1 warmup, or remove the Stage-3 classifier.}
  \label{fig:spice_ablation}
\end{figure}

Figure~\ref{fig:spice_ablation} shows that Full SPICE outperforms all controlled variants, validating the complete progressive design. Removing the Stage~1 warmup or supervised contrastive loss degrades performance, confirming that reliable visual-text learning and explicit POI-membership supervision are both necessary before structured groups can be generated. Adding a clustering fallback also weakens the result, showing that the group identifiers learned and generated by SPICE are more effective when used directly rather than revised by separate post-processing. Finally, removing the Stage~3 classifier causes the most pronounced degradation, demonstrating its central role in validating entries and assigning their semantic roles and completeness.

\subsection{POI Maintenance Evaluation}
\label{sec:poi_maintenance_eval}

\subsubsection{Quantitative Results} Table~\ref{tab:poi_maintenance} evaluates all methods on the same manually verified components and retrieved candidates, thereby isolating maintenance decisions from upstream extraction errors. Compared with the ML-based method, the strongest non-LLM baseline, RIMA with Qwen3.5-Plus raises Average F1 from 0.363 to 0.858 on Kuaishou Dashcam and from 0.535 to 0.880 on MAPbench. The consistent gains across $L1$--$L3$ show that visual and auxiliary evidence are crucial not only for retrieving plausible records, but also for resolving POI identity and determining database-attribute updates. Table~\ref{tab:poi_maintenance} also provides a controlled ablation of agent paradigms: One-Shot, ReAct, and RIMA use the same inputs, evidence sources, and backbone in each comparison. RIMA performs best on both datasets with all four backbones; under Qwen3.5-Plus, it exceeds the strongest alternative agent by 10.6 and 9.6 percentage points in Average F1 on Kuaishou Dashcam and MAPbench, respectively. RIMA also uses fewer calls and lower latency than ReAct. This consistent advantage validates RIMA's decision-before-attribute workflow, which resolves entity identity before attribute-level maintenance and invokes auxiliary evidence within a controlled process.

\subsubsection{Case Study}\label{sec:case_study} 
\begin{figure}[t]
  \centering
  \includegraphics[width=\columnwidth]{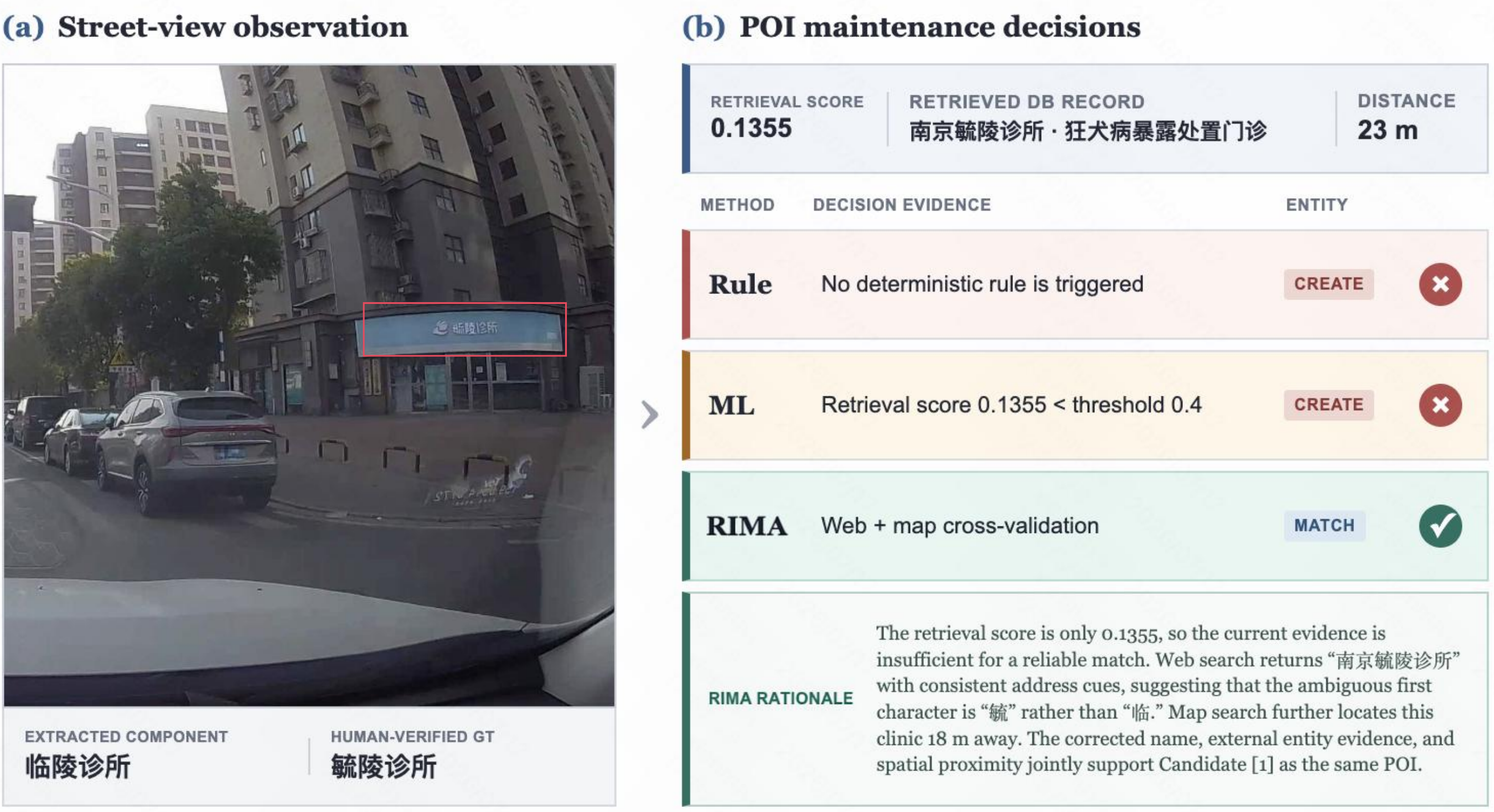}
  \caption{Low-confidence entity resolution: RIMA uses web and map evidence to match the existing POI, while both baselines recommend \texttt{create}.}
  \Description{A street-view image and three POI maintenance decisions. The perceived clinic name contains a character error. Rule-based and ML-based methods create a new record, while RIMA uses web and map evidence to match the existing clinic record.}
  \label{fig:case_study}
\end{figure}

Figure~\ref{fig:case_study} illustrates the mechanism behind these gains. A character error in the perceived POI name produces a low retrieval score, causing both Rule-based and ML-based methods to recommend \texttt{create} and introduce a duplicate record. RIMA instead treats the low-confidence match as unresolved, gathers complementary web and map evidence, and correctly links the observation to the existing POI. The case demonstrates how evidence escalation enables reliable identity resolution without globally relaxing retrieval thresholds.

\subsection{Street-view Image-based POI Data Production Evaluation}
\label{sec:full_task_eval}

Table~\ref{tab:full_task} evaluates the four extraction--maintenance combinations from raw images under the same database snapshot and human-review protocol. MAPS achieves the best recommendation quality and practical yield across all four metrics. The factorized comparison further demonstrates the complementarity between SPICE for POI extraction and RIMA for POI maintenance. MAPS's verified matches have a mean proxy distance of 38.6\,m, with 82.9\% within 50\,m, indicating that the image capture coordinates provide an adequate spatial proxy for this task; details are provided in Appendix~\ref{sec:appendix_distance}.

\begin{table}[!t]
  \caption{Human-reviewed full-task POI production under a fixed database snapshot.\#Accepted denotes accepted recommendations per 100 images; time is in s/image.}
  \label{tab:full_task}
  \centering
  \scriptsize
  \setlength{\tabcolsep}{2.5pt}
  \resizebox{\columnwidth}{!}{%
  \begin{tabular}{@{}llccccc@{}}
    \toprule
    \textbf{Extraction} & \textbf{Maintenance} &
    \textbf{Action} & \textbf{Attr.} & \textbf{\#Accepted} & \textbf{Comp.} & \textbf{Time} \\
    \midrule
    Qwen3.5-Plus & ML-based & 51.9 & 77.0 & 76.1 & 48.2 & 13.9 \\
    SPICE & ML-based & 45.8 & 75.2 & 68.5 & 43.4 & \textbf{4.65} \\
    Qwen3.5-Plus & RIMA & 69.6 & 92.9 & 104.6 & 66.2 & 38.2 \\
    \multicolumn{2}{l}{\textbf{MAPS (SPICE + RIMA)}} & \textbf{84.8} & \textbf{96.4} & \textbf{132.5} & \textbf{83.9} & 20.8 \\
    \bottomrule
  \end{tabular}%
  }
\end{table}

\section{SYSTEM DEPLOYMENT}
\label{sec:deployment}

\begin{figure}[b]
  \centering
  \includegraphics[width=\columnwidth]{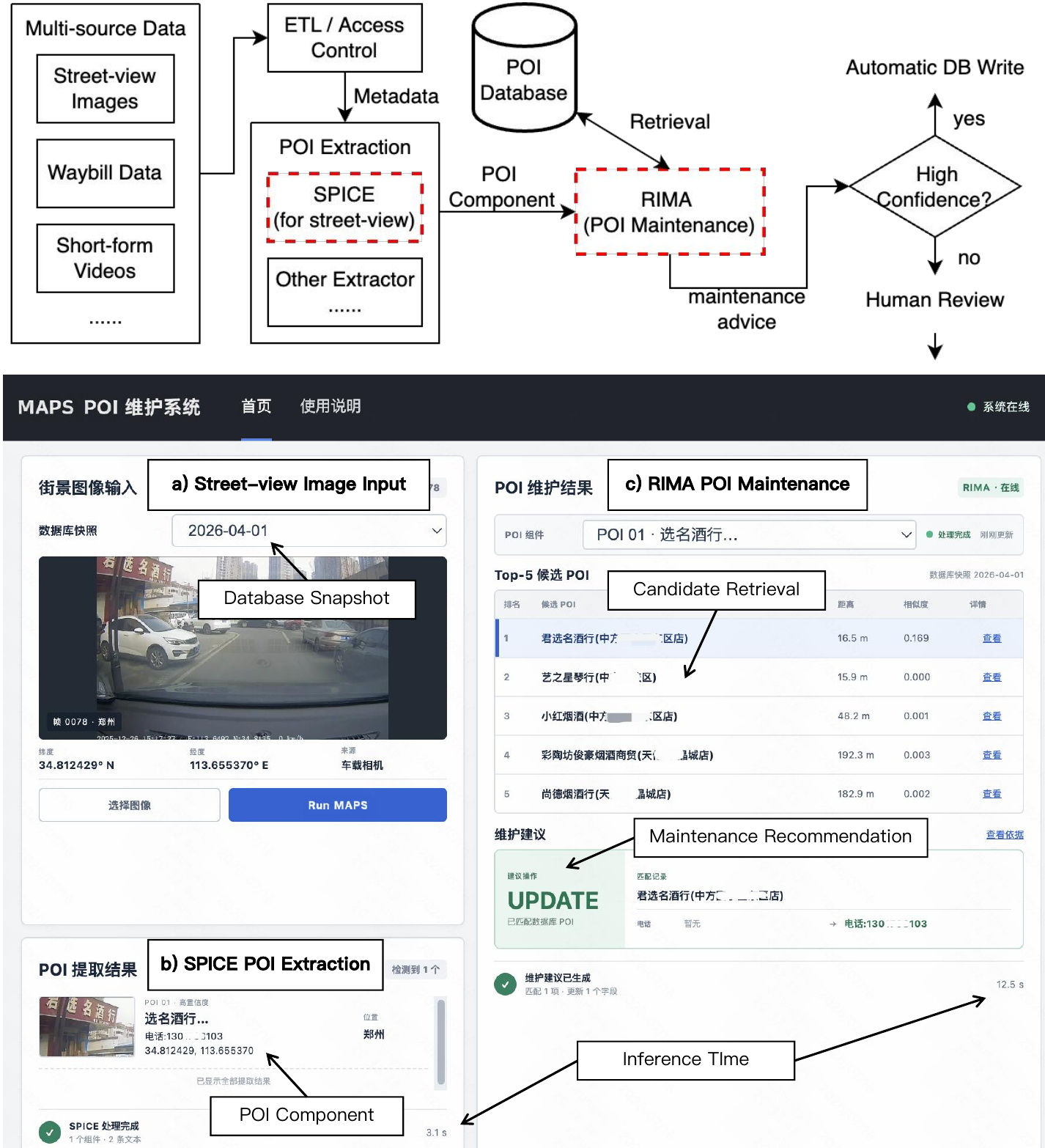}
  \caption{MAPS's deployed POI production pipeline and review interface; dashed boxes indicate deployed modules.}
  \label{fig:deployment}
\end{figure}

We have deployed MAPS in the POI production pipeline of Kuaishou, a leading short-video and live-streaming platform in China, as illustrated in Figure~\ref{fig:deployment}. An upstream ETL and access-control layer mitigates source-specific privacy, confidentiality, and commercial risks. SPICE converts geotagged street-view images into structured POI components, after which RIMA retrieves database candidates and generates attribute-level maintenance recommendations. High-confidence recommendations pass automated checks before database commit, whereas uncertain cases are routed to human review.

Kuaishou currently maintains tens of millions of candidate images, including a large collection of geotagged street-facing dashcam images covering millions of maintainable POIs. During the ongoing production trial, MAPS processes an image in 20.8s on average, compared with 90s for manual annotation. To date, MAPS has processed more than 100,000 images. The observed 76.9\% reduction corresponds to over 1,900 equivalent annotation hours saved. Extrapolating this reduction across the full collection suggests savings of hundreds of thousands of annotation hours. These maintained POIs support Kuaishou's local-service search and recommendation, merchant verification, and subsidy-fraud prevention.

\section{RELATED WORK}
\label{sec:related}
\noindent\textbf{Crowdsourced Spatio-temporal Data Mining.} Crowdsourced trajectories, delivery records, and street-level images provide low-cost but noisy observations of urban entities. Existing work recovers three information types. For locations, repeated GPS observations support road-network construction, while courier trajectories and delivery events reveal actual POI locations despite delayed or erroneous confirmations~\cite{chen2016city,yang2022duare,ruan2020doing,ruan2022discovering,waybillCoord2024}. For attributes, observations linked to roads or storefronts reveal traffic states, illegal parking, vehicle emissions, and storefront vacancy~\cite{qin2021network,he2018detecting,shang2014inferring,storefrontVacancy2024}. For names, delivery addresses provide POI names, and user-location patterns help identify aliases~\cite{waybillPoiCoord2021,he2021poi}. Most studies target one information type; MAPS instead connects structured POI extraction with database matching and attribute-level maintenance.

\noindent\textbf{POI Matching, Verification, and Database Updating.} POI conflation aligns heterogeneous records using attribute similarities and rule-based or learned classifiers~\cite{sun2023conflating}; pretrained language models cast entity matching as sequence-pair classification over serialized records~\cite{li2020deep}. Map services match incomplete queries to POIs using semantic and spatio-temporal context, auto-complete POIs, or extract POI--accessibility pairs from text~\cite{queryPoiMatching2020, poiAutoCompletion2021, gedit2021}. Industrial systems use multimodal street views to verify archived POIs or combine logistics data with crowd intelligence to detect new POIs and aliases~\cite{dumapper2022, urbanPoi2023}. These methods address record matching, search, verification, or update discovery separately. MAPS instead interacts with an existing POI database to jointly resolve POI identity and perform evidence-grounded attribute-level maintenance.

\section{CONCLUSION}
\label{sec:conclusion}

We study street-view image-based POI data production and present MAPS, which combines SPICE for extracting structured, database-ready POI components with RIMA for evidence-grounded identity resolution and attribute-level maintenance. 
Extensive experiments on real-world datasets have shown the superiority of each component of MAPS, and MAPS achieves a 21.8\% performance improvement over the strongest baseline.
MAPS has also been deployed in Kuaishou's production pipeline,  reducing average processing time by 76.9\% compared with manual annotation. One limitation is our reliance on storefront information visible from the street, which may cause MAPS to miss businesses without visible signage. We may broaden its coverage by incorporating other crowdsourced evidence, such as delivery records. In the future, the evidence-grounded removal of POIs from the database and applying MAPS to other crowdsourced data are worth to be investigated.



%

\bibliographystyle{ACM-Reference-Format}
\bibliography{reference}

\appendix

\section{Evaluation Protocols}
\label{app:evaluation_protocols}

\subsection{POI Extraction Metric Computation}
\label{sec:appendix_extraction_metrics}

POI extraction evaluation first checks whether an image contains any valid POI, then aligns predicted POI groups with ground-truth POI groups, and finally evaluates text extraction and role classification inside aligned groups. For any metric with counts $(\mathrm{TP},\mathrm{FP},\mathrm{FN})$, we use
\begin{equation}
  P=\frac{\mathrm{TP}}{\mathrm{TP}+\mathrm{FP}},\quad
  R=\frac{\mathrm{TP}}{\mathrm{TP}+\mathrm{FN}},\quad
  F1=\frac{2PR}{P+R}.
  \label{eq:appendix_prf}
\end{equation}

\noindent\textbf{Text normalization and similarity.}
For string-based matching, we normalize text by removing whitespace and common punctuation and applying case folding for Latin characters. Let $s(x,y)$ denote the SequenceMatcher similarity between normalized strings $x$ and $y$. Two strings are considered matched when
\begin{equation}
  s(x,y) \geq \tau,\qquad \tau=0.7 .
  \label{eq:appendix_text_sim}
\end{equation}
This threshold is used for both POI-name alignment and entry-level text matching unless otherwise specified.

\noindent\textbf{Has-POI F1.}
An image is positive if it contains at least one annotated valid POI. $F1_{\mathrm{hp}}$ is the binary F1 score over image-level has-POI predictions.

\noindent\textbf{POI-Name F1.}
For each image, let $G_{nm}$ and $P_{nm}$ be the ground-truth and predicted POI-name sets, excluding branch names. Define $\mu(x,S)=\mathbf{1}[\max_{y\in S}s(x,y)\geq\tau]$. Matching is performed bidirectionally:
\begin{equation}
  m_p=\sum_{p\in P_{nm}}\mu(p,G_{nm}),\qquad
  m_g=\sum_{g\in G_{nm}}\mu(g,P_{nm}).
\end{equation}
The POI-name counts are
\begin{equation}
  \mathrm{TP}_{\mathrm{nm}}=\min(m_p,m_g),\quad
  \mathrm{FP}_{\mathrm{nm}}=|P_{nm}|-m_p,\quad
  \mathrm{FN}_{\mathrm{nm}}=|G_{nm}|-m_g .
  \label{eq:appendix_nm_counts}
\end{equation}
This cap avoids duplicate credit when multiple predicted names match the same ground-truth name.

\noindent\textbf{Group-aware text extraction F1.}
Predicted and ground-truth POI groups are first aligned using POI-name matches. At least one matched POI-name entry is required for two groups to be treated as the same POI. Entry-level text matching is then performed only within aligned POI groups, using greedy one-to-one matching under Eq.~\eqref{eq:appendix_text_sim}. Let $M_e$ be the resulting one-to-one matched entry set, and let $N_g$ and $N_p$ be the total numbers of ground-truth and predicted entries. Then
\begin{equation}
  \mathrm{TP}_{\mathrm{txt}}=|M_e|,\quad
  \mathrm{FP}_{\mathrm{txt}}=N_p-|M_e|,\quad
  \mathrm{FN}_{\mathrm{txt}}=N_g-|M_e| .
  \label{eq:appendix_txt_counts}
\end{equation}
Entries in unmatched predicted POI groups therefore become false positives, while entries in unmatched ground-truth POI groups become false negatives. Non-POI texts are evaluated through flattened one-to-one text matching because they do not have a POI-group identity.

\noindent\textbf{Macro attribute-role F1.}
After entry-level matching, each matched pair contributes to a role confusion matrix over five roles. Let $\mathcal{R}=\{nm, br, addr, tel, \texttt{other}\}$ denote this role set. For each role $f\in\mathcal{R}$, we compute $F1_f$ from the role confusion matrix using Eq.~\eqref{eq:appendix_prf}, and report
\begin{equation}
  F1_{\mathrm{attr}}=\frac{1}{5}\sum_{f\in\mathcal{R}}F1_f .
  \label{eq:appendix_attr_f1}
\end{equation}
The Average F1 Score of POI Extraction Task is
\begin{equation}
  \mathrm{AvgF1}_{\mathrm{extract}}
  =
  \frac{
  F1_{\mathrm{hp}} +
  F1_{\mathrm{nm}} +
  F1_{\mathrm{txt}} +
  F1_{\mathrm{attr}}
  }{4}.
  \label{eq:extraction_average_f1}
\end{equation}

\subsection{POI Maintenance Metric Computation}
\label{sec:appendix_maintenance_metrics}

POI maintenance evaluation uses manually verified POI components and database candidates so that RIMA and baselines are compared under the same component-level evidence. The maintenance chain is evaluated at three levels.

\noindent\textbf{$L1$ routing.}
$L1$ evaluates entity-level routing between \texttt{match} and \texttt{create}. A predicted \texttt{match} is correct only when the matched database POI identifier agrees with the annotation. In the component-level POI maintenance evaluation, the manually verified candidate sets contain no ground-truth \texttt{discard} cases; therefore, a predicted \texttt{discard} is counted as an incorrect $L1$ outcome.
\begin{equation}
  L1=\frac{F1_{\mathrm{match}}+F1_{\mathrm{create}}}{2}.
  \label{eq:appendix_l1}
\end{equation}

\noindent\textbf{$L2$ update/skip.}
$L2$ is evaluated over ground-truth matched components along the complete decision chain. A prediction receives downstream credit only when $L1$ first selects the correct database POI and the system then correctly decides whether that record should be updated or skipped. An incorrect $L1$ decision is therefore retained as a downstream error rather than excluded from $L2$ evaluation.
\begin{equation}
  L2=\frac{F1_{\mathrm{update}}+F1_{\mathrm{skip}}}{2}.
  \label{eq:appendix_l2}
\end{equation}

\noindent\textbf{$L3$ attribute-level maintenance.}
$L3$ in the main table reports attribute-update quality F1 over the same matched-decision chain. An attribute-level prediction receives credit only when $L1$ selects the correct database POI; otherwise, the corresponding downstream attribute predictions remain errors rather than being removed from evaluation. Let $\mathcal{T}_{\mathrm{attr}}=\{nm, br, loc, cat, addr, tel\}$ denote the six evaluated attribute types. For each type $t\in\mathcal{T}_{\mathrm{attr}}$, the evaluator requires the proposed updated value to exactly match its ground-truth target value. A mismatched value contributes a false positive and leaves the corresponding ground-truth value as a false negative. Location predictions use weight 0.5, while all other attribute types use weight 1.0:
\begin{equation}
  \mathrm{TP}_{L3}=\sum_{t\in\mathcal{T}_{\mathrm{attr}}}w_t\mathrm{TP}_t,\quad
  \mathrm{FP}_{L3}=\sum_{t\in\mathcal{T}_{\mathrm{attr}}}w_t\mathrm{FP}_t,\quad
  \mathrm{FN}_{L3}=\sum_{t\in\mathcal{T}_{\mathrm{attr}}}w_t\mathrm{FN}_t.
  \label{eq:appendix_l3_counts}
\end{equation}
$L3$ is then the F1 computed from these weighted counts using Eq.~\eqref{eq:appendix_prf}. The Average F1 Score of POI Maintenance Task is
\begin{equation}
  \mathrm{AvgF1}_{\mathrm{maint}}=\frac{L1+L2+L3}{3}.
  \label{eq:maintenance_average_f1}
\end{equation}
Because $L2$ and $L3$ are evaluated over the matched-decision chain, errors at $L1$ propagate to downstream levels instead of being hidden by conditional attribute scoring.

\subsection{Full-task Metric and Anchor Matching}
\label{sec:appendix_full_task_protocol}

Full-task scoring first matches each generated recommendation to a manually reviewed POI component. We only compare components from the same source image. Within that image, names are normalized by removing whitespace, common punctuation, ellipses, and bracket variants. A generated recommendation is matched to a reviewed component when the normalized names are the same, one name contains the other, or the two names share enough common characters. Recommendations that cannot be matched to any reviewed component are treated as \texttt{discard}.

After this matching step, scoring uses the human-reviewed target of the matched component. For a predicted \texttt{match}, the matched database POI name and identifier must also agree with the human-reviewed target; otherwise the action is counted as incorrect even if the system chose \texttt{match}.

Action Accuracy is computed over \texttt{create}, \texttt{match}, and \texttt{discard}. We first map \texttt{update} and \texttt{skip} to \texttt{match}:
\begin{equation}
  \phi(a)=
  \begin{cases}
    \texttt{match}, & a\in\{\texttt{update},\texttt{skip}\},\\
    a, & a\in\{\texttt{create},\texttt{discard}\}.
  \end{cases}
  \label{eq:appendix_action_map}
\end{equation}
A matched recommendation is counted correct only if the normalized matched POI name also agrees with the human-reviewed target. Let
\begin{equation}
  c_i=\mathbf{1}[
  \phi(\hat{a}_i)=\phi(a_i)\wedge \mathrm{nmOK}_i],
  \label{eq:appendix_action_correct}
\end{equation}
then
\begin{equation}
  \mathrm{ActionAcc}=\frac{1}{|\mathcal{J}|}\sum_{i\in\mathcal{J}}c_i,
  \label{eq:appendix_action_acc}
\end{equation}
where $\mathcal{J}$ is the set of judged recommendations; $\mathrm{nmOK}_i$ is treated as true for non-match actions.

Attribute Accuracy is evaluated only on cases where both human and system actions are in $\{\texttt{update},\texttt{skip}\}$. Let $U_{\mathrm{ok}}$ be acceptable updates, $U_{\mathrm{bad}}$ unacceptable updates, $S_{\mathrm{ok}}$ correct skips, $N_{S\to U}$ unnecessary updates, and $N_{U\to S}$ missed updates. With $D_{\mathrm{attr}}=U_{\mathrm{ok}}+U_{\mathrm{bad}}+S_{\mathrm{ok}}+N_{S\to U}+N_{U\to S}$,
\begin{equation}
  \mathrm{AttrAcc}=\frac{U_{\mathrm{ok}}+S_{\mathrm{ok}}}{D_{\mathrm{attr}}}.
  \label{eq:appendix_attr_acc}
\end{equation}

Accepted Recommendations per 100 Images counts correct recommendations whose action type is \texttt{create}, \texttt{update}, or \texttt{skip}. Let $N_{\mathrm{ok}}$ denote the number of such accepted recommendations and $N_{\mathrm{img}}$ the number of evaluated images:
\begin{equation}
  \mathrm{Accepted}/100=\frac{N_{\mathrm{ok}}}{N_{\mathrm{img}}}\times 100 .
  \label{eq:appendix_accepted}
\end{equation}
Completion divides this yield by the manually reviewed upper bound $N_{\mathrm{human}}$, which counts human-accepted create, update, and skip recommendations:
\begin{equation}
  \mathrm{Completion}=\frac{N_{\mathrm{ok}}}{N_{\mathrm{human}}}.
  \label{eq:appendix_completion}
\end{equation}

\subsection{Reference Spatial Accuracy Analysis}
\label{sec:appendix_distance}

\begin{table}[!t]
  \caption{Reference spatial accuracy for manually verified correct matches with available coordinates.}
  \label{tab:appendix_distance}
  \centering
  \small
  \resizebox{\columnwidth}{!}{%
  \begin{tabular}{lrrrr}
    \toprule
    \textbf{System} & \makecell{\textbf{Valid}\\\textbf{matches}} & $\leq$50m & $\leq$100m & \makecell{\textbf{Mean}\\\textbf{dist.}} \\
    \midrule
    MAPS & 242 & 82.9\% & 92.2\% & 38.6m \\
    Qwen3.5-Plus + RIMA & 195  & 79.6\% & 91.0\% & 40.9m \\
    SPICE + ML-based & 126 & 78.9\% & 90.6\% & 41.2m \\
    Qwen3.5-Plus + ML-based  & 145 & 79.2\% & 91.9\% & 41.7m \\
    \bottomrule
  \end{tabular}
  }
\end{table}

We additionally compute a reference spatial accuracy statistic for matched recommendations in the full-task evaluation. This is not used as a main score because the evaluation data do not contain the camera calibration and pose information required for accurate visual geolocation, such as camera intrinsics, reliable extrinsics, depth, and a precise viewing ray for each observed storefront. Therefore, we cannot triangulate the exact POI coordinates from the image evidence alone. Instead, we use the vehicle-mounted image capture coordinate as a proxy observation location and compare it with the matched database POI coordinate. This proxy distance is also affected by the underlying map coordinates and shared retrieval service, rather than only by agent reasoning.

For a matched recommendation whose database POI identifier has been manually verified as correct and whose distance is available, we measure the geodesic distance $d_i$ between the image capture coordinate and the matched database POI coordinate. Recommendations with an incorrect matched POI identifier are excluded before computing the distance distribution. For a threshold $r$, the reported spatial ratio is
\begin{equation}
  \mathrm{Dist}(r)=\frac{1}{|\mathcal{D}|}\sum_{i\in\mathcal{D}}\mathbf{1}[d_i\leq r],
  \qquad r\in\{50\mathrm{m},100\mathrm{m}\},
  \label{eq:appendix_dist}
\end{equation}
where $\mathcal{D}$ is the set of manually verified correct-POI-identifier matches with available distances. Table~\ref{tab:appendix_distance} reports this reference analysis. MAPS obtains the highest share of matches within 50 meters while maintaining the strongest recommendation quality in the full-task evaluation (Table~\ref{tab:full_task}); its mean proxy distance is 38.6 meters, also below 50 meters. These values indicate that accepted matches are usually spatially close to the observed street-view evidence despite the absence of exact camera-geometry data, but they should be interpreted as a sanity check rather than as an independent measure of POI maintenance correctness.

\subsection{Human Review Criteria}
\label{sec:appendix_human_review}

Reviewers see the image, structured component, retrieved candidates, and any proposed matched record; when available, they may consult map, geocoding, offline verification, or web evidence. Review proceeds strictly from POI existence and identity to attribute changes. For component-level evaluation, a visually supported component is \texttt{match} if a candidate represents the same real-world POI and \texttt{create} otherwise; overly ambiguous components are excluded rather than labeled as positive \texttt{discard} cases. Matched components are then labeled \texttt{update}/\texttt{skip} with affected attributes.

For the full-task evaluation, each generated recommendation is compared with reviewed POIs from the same image using normalized-name agreement. Reviewers verify the action, matched identity, and attribute acceptability; unmatched recommendations are treated as \texttt{discard}. A wrong identity invalidates downstream attributes. Updates require evidence that improves the record, skips retain an adequate record, and creates require a real POI absent from the reviewed candidates.

\section{Data and Experimental Details}
\label{app:data_details}

\subsection{MAPbench Adaptation}
\label{sec:appendix_mapbench}

MAPbench serves two evaluation roles in this work. For zero-shot POI extraction evaluation, we use its street-view-related images as an external test source; none of these images are included in SPICE training or validation. To compute the same POI extraction metrics as on Kuaishou Dashcam and Meituan ReCTS, we normalize MAPbench annotations into our POI-entry format: POIs are treated as groups, while text that does not name a POI is not counted as an independent POI name. For POI maintenance evaluation, MAPbench also provides coordinate-grounded street-view components that can be paired with retrieved database candidates for maintenance evaluation. These adaptations are used only for evaluation and do not introduce MAPbench supervision during training.

\subsection{Training and Inference Settings}
\label{sec:appendix_hyperparams}

To obtain OCR proposals, we use PP-OCRv5 from PaddleOCR~\cite{ppocrv5_2026} to generate the text bounding boxes and initial transcriptions supplied to SPICE. SPICE uses Qwen3.5-0.8B as its backbone and is trained and evaluated locally on four 48GB GPUs. We split the 33,044-image training pool into training and validation subsets at a 9:1 ratio. The validation subset is used only for checkpoint selection and early stopping; neither the target-domain test sets nor MAPbench is used for model selection. We use a random seed of 42 throughout. Stage~1 trains rank-64 LoRA for up to 3 epochs with a per-device batch size of 2, 64 gradient-accumulation steps, and a learning rate of $1.5\times10^{-4}$. Stage~2 performs full-parameter fine-tuning for up to 5 epochs with a per-device batch size of 2, 16 accumulation steps, and a learning rate of $2\times10^{-5}$. Both stages use AdamW, weight decay 0.01, cosine learning-rate decay with a 0.03 warmup ratio, bfloat16 precision, and a maximum sequence length of 4096. Stage~3 freezes the Stage-2 backbone and trains the classifier with a batch size of 16 and a learning rate of $10^{-4}$ for at most 50 epochs, using early stopping with a patience of 10. In our implementation, the hidden, grouping, and text representation dimensions are $d_{\mathrm{hid}}=1024$, $d_{\mathrm{grp}}=128$, and $d_{\mathrm{txt}}=1024$, respectively. For all three stages, we select the checkpoint with the lowest loss on the validation subset.

For RIMA and VLM-agent baselines, all LLM/VLM calls use temperature 0. Retrieval uses the same fixed top-$K$ candidate budget with $K{=}5$ for all compared methods. ReAct baselines are allowed at most 10 tool-interaction steps. Time is measured as average seconds per image for POI extraction, average seconds per component for POI maintenance, and image-to-recommendation latency for the full-task evaluation.

\section{Method and System Details}
\label{app:method_details}

\subsection{SPICE Construction and Grouping Details}
\label{app:spice_details}

The association between OCR proposals and reconstructed entries is not constrained to be one-to-one. One proposal may correspond to a complete entry, several adjacent proposals may be merged, and a visually supported entry may be recovered without a proposal association. For associated proposals, their union region forms the entry box $\beta_i$ used by the position view. A recovered entry without a reliable association remains available as text but has no reliable box.

The supervised contrastive set is constructed independently within each image. Here, $g_i$ denotes the ground-truth group identifier of entry $i$, whereas $\hat g_i$ is the identifier generated by the LM head. Let $\mathcal{I}$ be the indices of entries whose $g_i$ is positive, and let $P(i)=\{j\in\mathcal{I}\setminus\{i\}:g_j=g_i\}$ contain the other entries assigned to the same ground-truth POI as anchor $i$. Anchors without a positive partner are omitted; their remaining index set is $\mathcal{I}_{+}=\{i\in\mathcal{I}:|P(i)|>0\}$. The loss is
\begin{equation}
  \mathcal{L}_{\mathrm{SupCon}}
  =-\frac{1}{|\mathcal{I}_{+}|}
  \sum_{i\in\mathcal{I}_{+}}\frac{1}{|P(i)|}
  \sum_{j\in P(i)}
  \log
  \frac{\exp((h_i^{\mathrm{grp}})^\top h_j^{\mathrm{grp}}/\tau)}
  {\sum_{u\in\mathcal{I}\setminus\{i\}}\exp((h_i^{\mathrm{grp}})^\top h_u^{\mathrm{grp}}/\tau)}.
  \label{eq:supcon}
\end{equation}
Here, $i$ is an anchor entry, $j$ indexes one of its positive entries in $P(i)$, and $u$ indexes any non-anchor entry in $\mathcal{I}$. The quantities $|P(i)|$ and $|\mathcal{I}_{+}|$ are the numbers of positive entries for anchor $i$ and valid anchors in the image, respectively, while $\tau$ is the temperature. Since $h_i^{\mathrm{grp}}$ is normalized, the inner product is cosine similarity. Entries with $g_i=0$ are excluded because they do not define a POI identity. The grouping head outputs a 128-dimensional representation. In Eq.~\eqref{eq:spg_loss}, we set $\lambda_g=0.3$ and $\lambda_e=1.5$.

\subsection{Multi-view Classifier Details}
\label{app:classifier_details}

The hidden view $h_i^{\mathrm{hid}}$ is the final backbone state associated with entry $i$ and retains its image and generated-text context. The grouping view $h_i^{\mathrm{grp}}$ is the normalized output of the Stage-2 grouping head. The text view $h_i^{\mathrm{txt}}$ is obtained by mean-pooling the adapted backbone's token embeddings of $\hat t_i$. The position view is
\begin{equation}
  h_i^{\mathrm{pos}}=\operatorname{Geom}(\beta_i),
  \label{eq:position_view}
\end{equation}
where $\operatorname{Geom}(\cdot)$ extracts the normalized center, width, height, and minimum horizontal and vertical distances from the entry box $\beta_i$ to the image boundary. Thus, $h_i^{\mathrm{pos}}\in\mathbb{R}^{6}$. The boundary distances provide cues for truncated text, and an entry without a reliable proposal association receives a zero position vector.

In our implementation, $d_{\mathrm{hid}}=1024$, $d_{\mathrm{grp}}=128$, and $d_{\mathrm{txt}}=1024$. The projected views enter a 512-dimensional, three-layer Transformer followed by separate validity, role, and completeness heads. The classifier objective is
\begin{equation}
  \mathcal{L}_{\mathrm{cls}}
  =\lambda_v\mathcal{L}_{\mathrm{valid}}
  +\lambda_r\mathcal{L}_{\mathrm{role}}
  +\lambda_q\mathcal{L}_{\mathrm{complete}},
  \label{eq:cls_loss}
\end{equation}
Here, $\mathcal{L}_{\mathrm{cls}}$ is the total classifier loss. The terms $\mathcal{L}_{\mathrm{valid}}$, $\mathcal{L}_{\mathrm{role}}$, and $\mathcal{L}_{\mathrm{complete}}$ are the cross-entropy losses for validity, semantic role, and completeness, respectively; $\lambda_v$, $\lambda_r$, and $\lambda_q$ are their weights. We use $\lambda_v=1.0$, $\lambda_r=2.0$, and $\lambda_q=1.0$, with label smoothing 0.05 and class weighting for role and completeness.

\subsection{RIMA Evidence Escalation}
\label{app:rima_details}

RIMA requests additional evidence only when the component and retrieved records do not support a clear decision. Typical triggers include several nearby branches with similar names, an incomplete storefront name, disagreement between name similarity and geographic distance, or conflicting business and category information. Map or geocoding evidence checks spatial correspondence; candidate-conditioned verification of the retained street-view image checks storefront identity; category knowledge resolves ambiguous business types; and entity or web search supplies missing branch information. The evidence is returned to the same evidence-grounded decision and does not change the \texttt{match}/\texttt{create}/\texttt{discard} space or the decision-before-attribute order.

\subsection{Attribute-level Reasoning and Recommendation Schema}
\label{app:attribute_details}

In attribute-level reasoning, RIMA evaluates the name, branch name, phone number, location, address, and category using the component and retained evidence. Category decisions use business semantics in auxiliary text and image evidence, consulting category knowledge when needed. Each attribute is assessed independently to determine whether the evidence supports adding or changing its value, retaining the database value, or omitting an unreliable observation.

For a \texttt{match}, any supported attribute update yields $decision=\texttt{update}$, and $op$ contains only the supported changes; otherwise, $decision=\texttt{skip}$ and $op$ is empty. For \texttt{create}, $decision=\texttt{create}$ and $op$ retains only validated attributes; for \texttt{discard}, $decision=\texttt{discard}$ and $op$ is empty. The final action $act$ retains the resulting $decision$ and $op$, the optional matched record, rationale, and supporting evidence.

\section{Ethics and Privacy Statement}
\label{app:ethics_privacy}

We study POI database maintenance using street-view imagery, public benchmarks, and an internal POI database snapshot. We focus on storefronts and business records rather than individuals; we neither use personal identities nor infer sensitive personal attributes. We use only street-view imagery in which people, vehicles, faces, license plates, and other incidental content have been redacted; such content remains outside our annotation and evaluation scope. We use the internal imagery and database records only for research and system evaluation under organizational data-governance requirements and do not release them publicly. In deployment, we apply confidence-gated database updates: high-confidence results pass production checks before automatic commit, while low-confidence recommendations require human verification. Nevertheless, OCR and entity-resolution errors could lead to incorrect create or update recommendations; we mitigate this risk through conservative \texttt{skip}/\texttt{discard} actions, confidence routing, and manual review of uncertain cases. Performance may also vary across regions, languages, storefront conditions, and business categories, so we do not claim uniform generalization beyond the evaluated settings.

\end{document}